\documentclass[letterpaper, 10 pt, conference]{ieeeconf}
\IEEEoverridecommandlockouts
\usepackage{times}
\usepackage{cuted}
\usepackage{capt-of}
\usepackage{multicol}
\usepackage{graphicx}   %
\usepackage[sort,compress]{cite} 
\newcommand{\citep}{\cite}       

\usepackage{amsmath}
\usepackage{amssymb}
\usepackage{cases}
\usepackage{graphicx}
\usepackage[sort,compress]{cite}
\usepackage{booktabs}

\usepackage{amsthm}

\theoremstyle{definition}

\newtheorem{problem}{Problem}

\usepackage{mdframed} 

\usepackage[table]{xcolor}
\usepackage{subcaption}
\usepackage{wrapfig}
\usepackage{multirow}
\usepackage{mathtools}
\usepackage{empheq}
\usepackage{color}
\usepackage{bbm}
\usepackage{xspace}
\let\labelindent\relax
\usepackage[shortlabels]{enumitem}
\usepackage{tabularray}
\usepackage[final]{microtype} 

\usepackage{tikz} 
\usepackage{tkz-graph}
\usepackage{tkz-berge}
\usetikzlibrary{backgrounds,fit,shapes,snakes,arrows,shapes.geometric,positioning}
\usetikzlibrary{intersections,patterns,shapes.misc}
\usetikzlibrary{decorations.pathmorphing}
\usepackage{pgfplots}

\tikzstyle{block} = [rectangle, rounded corners, minimum width=3cm, minimum height=1cm,text centered, draw=black, fill=red!30]
\tikzstyle{new} = [rectangle, rounded corners, minimum width=1cm, minimum
height=1cm,text centered, draw=black, fill=blue!10!white, dashed]
\tikzstyle{arrow} = [thick,->,>=stealth]
\usetikzlibrary{calc, quotes}

\tikzstyle{fblock} = [rectangle, draw, fill=gray!20, 
text width=8em, text centered, rounded corners, minimum height=4em, minimum width = 8em]
\tikzstyle{line} = [draw, -latex']

\usepackage{algorithmicx}
\usepackage{algorithm}
\usepackage[noend]{algpseudocode}

\usepackage{fontawesome}

\usepackage{soul}
\usepackage{comment}
\usepackage[]{url}
\usepackage[font=small]{caption}
\usepackage[normalem]{ulem}
\useunder{\uline}{\ul}{}

\usepackage[colorlinks=true, linkcolor=blue, citecolor = black, urlcolor = cyan, filecolor=black, pagebackref=false, hypertexnames=false]{hyperref}
\usepackage{cleveref}
\crefname{problem}{Problem}{Problems}
\crefname{example}{Example}{Examples}
\crefname{section}{Sec.}{Secs.}
\Crefname{section}{Section}{Sections}
\Crefname{table}{Table}{Tables}
\crefname{table}{Table}{Tabs.}
\crefname{figure}{Fig.}{Figs.}
\crefname{algorithm}{Algorithm}{Algorithms}
\crefname{remark}{Remark}{Remarks}
\crefname{theorem}{Theorem}{Theorems}
\crefname{proposition}{Proposition}{Propositions}
\crefname{lemma}{Lemma}{Lemmas}
\crefname{corollary}{Corollary}{Corollaries}
\crefname{assumption}{Assumption}{Assumptions}
\crefname{definition}{Definition}{Definitions}
\newcommand{\AlgName}{{\textsf{SBP}}\xspace}

\newcommand{\bmat}{\begin{bmatrix}}
\newcommand{\emat}{\end{bmatrix}}

\newcommand{\eg}{\emph{e.g.}\xspace}
\newcommand{\ie}{\emph{i.e.}\xspace}

\newcommand{\myParagraph}[1]{\vspace{0.5ex}{\bf #1.}\xspace}

\providecommand{\optional}[1]{{}}

\providecommand{\techreport}[1]{{}}  

\newcommand{\calD}{\mathcal{D}}

\newcommand{\calF}{\mathcal{F}}
\newcommand{\calG}{\mathcal{G}}

\newcommand{\calL}{\mathcal{L}}

\newcommand{\calX}{\mathcal{X}}
\newcommand{\calY}{\mathcal{Y}}

\title{\LARGE \bf Seeing the Bigger Picture: \\ 3D Latent Mapping for Mobile Manipulation Policy Learning
}

\author{
  Sunghwan Kim \qquad
  Woojeh Chung \qquad
  Zhirui Dai \qquad
  Dwait Bhatt \qquad
  Arth Shukla \\
  Hao Su \qquad
  Yulun Tian \qquad
  Nikolay Atanasov
}

\begin{document}
\bstctlcite{IEEEexample:BSTcontrol}
\maketitle
\begin{strip}\centering
    \vspace{-12mm}
    \includegraphics[trim=0 0 0 5, clip, width=\linewidth]{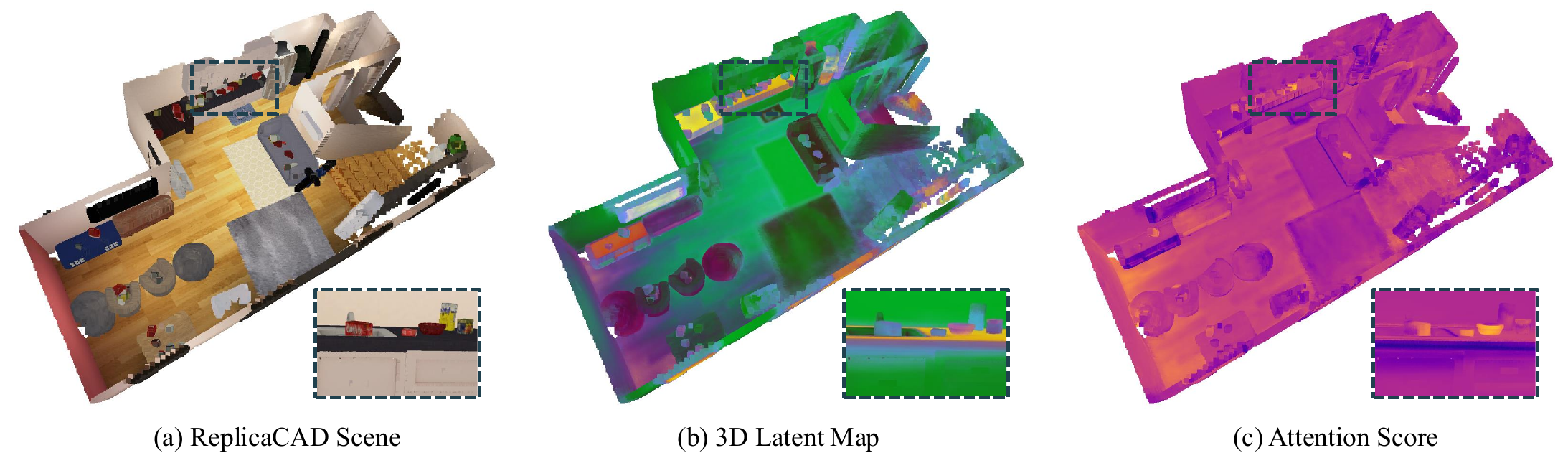}
    {\captionsetup{hypcap=false}
    \captionof{figure}{(a) An RGB rendering of a ReplicaCAD~\cite{szot2021habitat} scene. (b) A 3D latent feature map constructed by our method, visualized with PCA. (c) Attention weights on the latent map during task execution (\eg, ``pick up the bowl''), highlighting regions attended by the policy model.}\label{fig:teaser}
    }
\end{strip}

\begingroup
\def\thefootnote{\fnsymbol{footnote}}
\makeatletter\def\Hy@Warning#1{}\makeatother
\footnotetext{We gratefully acknowledge support from NSF CCF-2402689 (ExpandAI), ONR N00014-23-1-2353, and the Technology Innovation Program (20018112, Development of autonomous manipulation and gripping technology using imitation learning based on visual and tactile sensing) funded by the Ministry of Trade, Industry \& Energy (MOTIE), Korea.}
\footnotetext{S. Kim, Z. Dai, D. Bhatt, and N. Atanasov are with the Department of Electrical and Computer Engineering, University of California San Diego, La Jolla, CA 92093, USA, e-mail: \{suk063, zhdai, dhbhatt, natanasov\}@ucsd.edu.}
\footnotetext{W. Chung, A. Shukla, and H. Su are with the Department of Computer Science and Engineering, University of California San Diego, La Jolla, CA 92093, USA, e-mail: \{w5chung, arshukla, haosu\}@ucsd.edu.}
\footnotetext{Y. Tian is with the Robotics Department, University of Michigan, Ann Arbor, MI 48109, USA, e-mail: yulunt@umich.edu.}
\endgroup

\begin{abstract}
In this paper, we demonstrate that mobile manipulation policies utilizing a \emph{3D latent map} achieve stronger spatial and temporal reasoning than policies relying solely on images.
We introduce \emph{Seeing the Bigger Picture} (\AlgName), an end-to-end policy learning approach that operates directly on a 3D map of latent features. 
In \AlgName, the map extends perception beyond the robot's current field of view and aggregates observations over long horizons.
Our mapping approach incrementally fuses multiview observations into a grid of scene-specific latent features. 
A pre-trained, scene-agnostic decoder reconstructs target embeddings from these features and enables online optimization of the map features during task execution.
A policy, trainable with behavior cloning or reinforcement learning, treats the latent map as a state variable and uses global context from the map obtained via a 3D feature aggregator.
We evaluate \AlgName on scene-level mobile manipulation and sequential tabletop manipulation tasks.
Our experiments demonstrate that \AlgName (i) reasons globally over the scene, (ii) leverages the map as long-horizon memory, and (iii) outperforms image-based policies in both in-distribution and novel scenes, \eg, improving the success rate by 15\% for the sequential manipulation task.
\end{abstract}


\IEEEpeerreviewmaketitle

\section{Introduction}
\label{sec:introduction}

Recent advances in robot learning have led to remarkable progress in manipulation in semi-structured environments \cite{black2024vision, chi2023diffusion, zhao2023learning}. 
State-of-the-art systems use large vision–language models (VLMs), whose rich semantic priors and cross-modal reasoning translate natural language commands directly into low-level actions \cite{black2024vision}.
The next frontier lies in extending these methods beyond fixed tabletop setups to long-term mobile manipulation at room, building, neighborhood, and even larger scales.
However, existing methods rely on \emph{2D image-based} designs that operate directly on raw video streams.
While effective for short-term action prediction, image-based approaches struggle with consistent 3D understanding and long-horizon reasoning---two capabilities critical for spatially or temporally extended tasks.

In this work, we advocate for an alternative \emph{3D map-based} design that conditions robot policy learning on an explicit 3D representation of the environment.
A growing body of recent work explores 3D scene representations for manipulation. Some methods lift 2D foundation-model features into 3D on a per-frame basis \cite{gervet2023act3d, ke20243d, ze2023gnfactor, wang2024gendp}.
Another line of work encodes raw point cloud observations directly with specialized 3D backbones \cite{ze20243d, yang2024equibot}.
While both families preserve geometry and enhance local scene understanding, reconstructing the scene from scratch at each time step compromises temporal consistency and hinders long-horizon reasoning.
Complementary efforts fuse multiview observations into feature fields offline \cite{rashid2023language, shen2023distilled, wang2023d}.
Although these feature fields improve multiview consistency, they are confined to tabletop setups where the entire workspace remains visible at every time step and cannot adapt on the fly to novel views.

A key idea in this paper is to condition the robot policy on global features obtained from a persistent 3D map (see \cref{fig:teaser}), built incrementally from streaming observations.
Persistent maps have long benefited robot navigation \cite{durrant2006simultaneous, cadena2016past}, yet their potential to enhance manipulation remains under-explored.
Such maps offer two key advantages for mobile manipulation.
First, they act as \emph{spatial memory}, offering global visibility of object locations and task goals while mitigating occlusions from the current field of view.
Second, by accumulating information over time, they provide \emph{long-term context} that enables policies to reason beyond short observation windows.
\cref{fig:teaser}{\color{blue}b} shows a latent map containing spatially grounded language features, generated by our method.
\cref{fig:teaser}{\color{blue}c} illustrates that a policy conditioned on the map may attend to the entire scene, leveraging spatially and temporally extended context.

    
\textbf{Contributions.} We present \emph{Seeing the Bigger Picture} (\AlgName), an end-to-end policy learning method operating directly on an incrementally constructed 3D latent feature map.
We propose a modular design that comprises (i) a \emph{scene-specific} latent feature grid that aggregates and compresses multiview visual observations, and (ii) a \emph{pre-trained}, \emph{scene-agnostic} decoder that reconstructs target embeddings (\eg, CLIP \cite{radford2021learning}) from the latent features and generalizes to diverse scenes.
During task execution, our approach enables online optimization of latent features by using the pre-trained decoder.
To harness the 3D latent map in mobile manipulation tasks, we propose a 3D feature aggregator that summarizes the spatially distributed features into a compact \emph{global map token} that provides global context to support robot policy learning.
The global map token can be integrated into a behavior cloning or reinforcement learning policy, and we show that the resulting policy improves long-horizon reasoning and task execution.
Our contributions are summarized as follows.
\begin{itemize}
    \item We propose a mapping approach that incrementally builds a 3D map of latent features. Its modular design decouples scene-specific feature optimization from a scene-agnostic feature decoder, enabling generalization across different environments.
    
    \item We design a policy that treats the map as a state and tokenizes its features with a 3D feature aggregator to improve spatial and temporal reasoning. The model supports both behavior cloning and reinforcement learning.
    
    \item We demonstrate that our \AlgName method reasons globally and leverages the map as spatiotemporal memory in long-horizon manipulation tasks, outperforming image-based policies in both in-distribution and novel scenes.
\end{itemize}

\section{Related Work}
\label{sec:related}

\subsection{3D Feature Mapping with Large Vision Models}
A growing body of work distills vision model features, \eg, CLIP~\cite{radford2021learning} and DINOv2~\cite{oquab2023dinov2}, into neural scene representations for grounding semantics in 3D. 
Early work such as LERF~\cite{kerr2023lerf} embeds vision-language features into implicit neural fields by enforcing multiview consistency, enabling open-vocabulary queries.
To achieve fine-grained segmentation, subsequent works adopt explicit 3D representations (\eg, Gaussian splatting~\cite{kerbl20233d}) and leverage SAM~\cite{kirillov2023segment} to obtain precise object boundaries~\cite{qin2024langsplat,qiu2024feature}.
Recent work extends these representations to dynamic environments via online updates~\cite{katragadda2025online} and 4D scene representations for moving objects~\cite{yang2023emernerf,li20254d}.

Several works structure the 3D features to support downstream manipulation tasks, such as 6-DoF grasping. 
F3RM~\cite{shen2023distilled} distills CLIP features into a hierarchical 3D feature grid, enabling few-shot grasping by optimizing grasp poses based on language queries. 
LERF-TOGO~\cite{rashid2023language} addresses non-uniform activations in LERF~\cite{kerr2023lerf} via DINO-based masking, improving part-aware grasp selection. 
GeFF~\cite{qiu2024learning} eliminates the need for per-scene optimization by introducing a generalizable NeRF~\cite{mildenhall2021nerf} encoder that predicts language-aligned features. 
OK-Robot~\cite{liu2024ok} presents a modular pipeline for language-conditioned household manipulation, while DynaMem~\cite{liu2024dynamem} utilizes a dynamic 3D voxel map with semantic vectors for object localization and navigation.
More recently, UAD~\cite{tang2025uad} distills visual affordance maps from VLMs into a task-conditioned predictor that generalizes in the wild.

Despite their strong 3D grounding capabilities, these methods focus on localizing objects or predicting affordances from language queries rather than leveraging the representations for end-to-end policy learning.
To bridge this gap, we introduce a 3D latent mapping approach designed to facilitate policy training for long-horizon tasks.

\subsection{Manipulation Policy Learning with 3D Representations}
3D scene representations encode explicit geometric structure (\eg, free space, surfaces) that is absent from 2D images, providing a stronger spatial inductive bias for manipulation. As a result, conditioning manipulation policies on 3D representations improves generalization and sample efficiency.

One line of work lifts features from 2D large vision models into 3D structures.
PerAct~\cite{shridhar2023perceiver} encodes RGB-D observations into voxel grids and predicts end-effector poses via a transformer.
Act3D~\cite{gervet2023act3d} lifts multi-scale CLIP features into a 3D feature cloud and scores candidate points via cross-attention to regress manipulator poses. 
This has been extended to generate action trajectories by conditioning diffusion models on the featurized 3D scene \cite{ke20243d}.
GNFactor \cite{ze2023gnfactor} trains a neural feature field that jointly reconstructs 2D foundation-model features and predicts robot actions in 3D.

Another line of work bypasses 2D feature lifting and learns policies directly from geometric observations like point clouds.
DP3 encodes raw point clouds into compact embeddings that condition a diffusion policy for manipulation~\cite{chi2023diffusion,ze20243d}.
Some approaches render the scene and apply multi-stage transformers that zoom in on regions of interest~\cite{goyal2024rvt2}.
Recent work fuses point clouds with DINO features, forming semantic fields that enable part-level generalization across object instances~\cite{wang2024gendp}.
EquiBot~\cite{yang2024equibot} combines $\mathrm{SIM}(3)$-equivariant architectures with a point cloud diffusion policy for scale, rotation, and translation invariance, while ActionFlow~\cite{funk2024actionflow} leverages $\mathrm{SE}(3)$-invariant point relations with flow matching for symmetry-preserving trajectories.

Despite recent progress, most approaches treat 3D representations as instantaneous observations, reconstructing them at each step.
We address this by learning end-to-end policies conditioned on a persistent 3D latent map, enabling global and long-horizon reasoning.

\section{Problem Formulation} \label{sec:problem}

This paper has two primary objectives: first, to incrementally construct a 3D latent feature map of a robot's workspace (\cref{sec:latent_prob}), and second, to design manipulation policies that use the map as a state variable to execute tasks specified in natural language (\cref{sec:manipulation_prob}).

\subsection{Latent Feature Mapping}
\label{sec:latent_prob}
Let $\calX\subseteq\mathbb{R}^{3}$ denote the robot's workspace and $\calY\subseteq\mathbb{R}^{k}$ a target embedding space (\eg, of language features such as CLIP \cite{radford2021learning}). 
We seek an efficient representation $m: \calX \to \calY$ that maps 3D workspace points to target embeddings.

\begin{problem}
\label{pro:mapping}
Given a dataset of coordinate-feature pairs $\mathcal{D}\!=\!\{(x, y)\}\subset\calX\times\calY$, find a map $m$ that minimizes the reconstruction loss:
\begin{equation}
\label{eq:mapping}
    \min_{m}\;
\mathbb{E}_{(x,y)\sim\mathcal{D}}
\bigl[
\calL\bigl(m(x),y\bigr)
\bigr],
\end{equation}
where $\calL:\calY\times\calY\to\mathbb{R}_{\ge 0}$ is a distance function on $\calY$.
\end{problem}

\begin{figure}[t]
  \centering
 \includegraphics[width=0.9\linewidth]{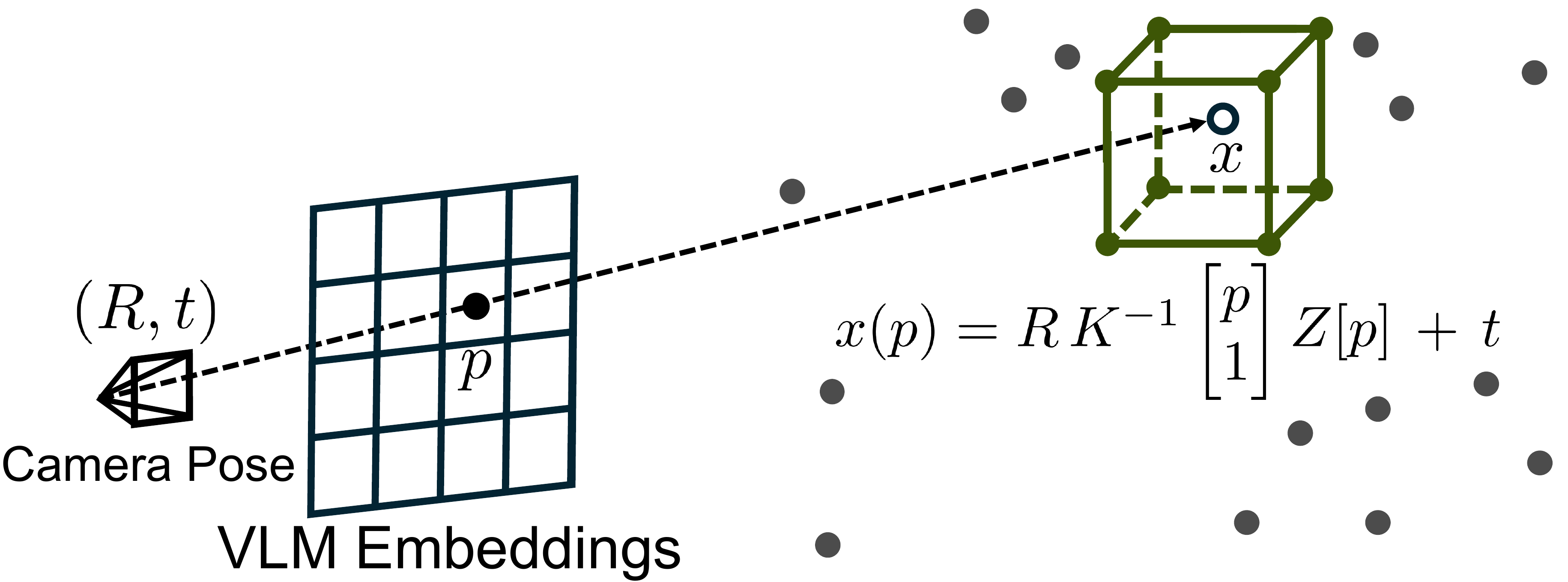}
  \caption{Visualization of per-patch VLM embeddings back-projected into the 3D world frame using depth $Z[p]$ and camera pose $(R,t)$. }
  \label{fig:backproject}
\end{figure}

To instantiate \cref{pro:mapping} for learning language-grounded latent maps, we use dense visual features extracted from a VLM as the target labels $y$. Given an RGB image $o$, a depth image $Z$, and camera pose $(R, t)$, we compute per‐patch embeddings $G \in \mathbb{R}^{k \times (H \times W)}$ by feeding the image through the VLM’s vision encoder. 
Following the ViT \cite{dosovitskiy2020image} convention, we partition the image into $H \times W$ non-overlapping patches, each producing a $k$-dimensional feature embedding at its corresponding spatial location.
We back-project each patch $p$ with valid depth $Z[p]$ into the world frame using the camera intrinsics $K$ and pose $(R,t)$, as shown in \cref{fig:backproject}: 
\begin{equation}
\label{eq:backproj}
    x(p) = R\,K^{-1}
    \begin{bmatrix} p \\ 1 \end{bmatrix}
    Z[p] \, + \, t.
\end{equation}
Each 3D point $x(p)$ is paired with its embedding $y(p)=G[p]$ from the VLM's vision encoder, yielding a coordinate-feature pair $(x,y)$. 
By aggregating these pairs across multiple viewpoints, we construct a training set $\calD$.
Minimizing \eqref{eq:mapping} ensures that the latent map captures the semantics provided by the VLM and associates them with 3D spatial locations.

\subsection{Map-Conditioned Policy Learning}
\label{sec:manipulation_prob}

We consider a manipulator that executes language-specified instructions. The task specification is embedded in a vector $e_\ell$ described below. At each time step $\tau$, the robot receives its proprioceptive state $s_\tau$ (\eg, joint angles) and observations $o_\tau$ (\eg, RGB images). 
A key feature of our formulation is the 3D latent map $m_\tau$, which is built incrementally from the observations and serves as persistent spatial memory of the environment (see \cref{fig:teaser}).
A policy $\pi_\phi$ with parameters $\phi$ maps $m_\tau$, state $s_\tau$, the observation $o_\tau$, and the task embedding $e_\ell$ to an action $a_\tau$ (\eg, joint velocities).

The policy is trained in two settings.
In behavior cloning (BC)~\cite{pomerleau1989alvinn}, the policy learns to mimic expert actions from a demonstration dataset~$\mathcal{D}^*$. For BC, we use a text embedding of a command (\eg, ``pick up the bowl'') as the task embedding~$e_\ell$.
In reinforcement learning (RL), the policy is trained to maximize expected sum of discounted rewards, $\mathbb{E}_{\mathcal{T} \sim \pi_\phi} [ \sum_\tau \gamma^\tau R_\tau ]$. The reward $R_\tau$ is shaped to facilitate learning, with bonuses for completing subgoals such as reaching, grasping, or placing an object. For RL, we use a learnable task embedding specifying the target object \cite{hansen2023td}.

\begin{problem}
\label{pro:policy}
Determine policy parameters $\phi$ that enable the robot to complete its task by solving $\min_{\phi}\; \mathcal{J}(\pi_\phi)$,
where the form of $\mathcal{J}(\pi_\phi)$ depends on the learning method:
\begin{equation}
\label{eq:rl_bc_objectives}
\mathcal{J}(\pi_\phi) \!=\!
\begin{cases} 
    -\mathbb{E}_{(\cdot, a^*_\tau) \sim \mathcal{D}^*} \left[ \log \pi_\phi(a^*_\tau \mid m_\tau, o_\tau, s_\tau, e_\ell) \right] & \text{(BC)}
    \\[1.5ex]
    -\mathbb{E}_{\mathcal{T} \sim \pi_\phi} \left[ \sum_\tau \gamma^\tau R_\tau \right] & \text{(RL)}
\end{cases}
\end{equation}
\end{problem}

\section{Latent Feature Mapping}
\label{sec:mapping}

In this section, we present our latent mapping approach, grounded in two principles. 
(1)~\emph{Incremental updates}: A 3D latent map continuously integrates multiview observations to serve as persistent spatial memory. 
(2)~\emph{Modularity}: A scene-specific representation is decoupled from a scene-agnostic mapping function to enable generalization across scenes.

To realize this modularity, we model the latent map as an encoder-decoder architecture, $m{=}(F_{\psi},D_{\theta})$. The encoder $F_{\psi}\!:\!\calX\to\calF$ lifts workspace points $x\in\calX$ to a latent space $\calF$, and the decoder $D_{\theta}\!:\!\calF\to\calY$ projects latent features $f \in \calF$ to the target space $\calY$.
The intermediate space $\calF$ enables the map to capture the geometric and semantic structure of the environment more effectively than a direct $\calX\to\calY$ mapping~\cite{muller2022instant, takikawa2021neural, kerr2023lerf}.
This architecture disentangles the scene-specific encoder parameters $\psi$ from the scene-agnostic decoder parameters $\theta$. 
Pre-training the decoder on diverse environments enables fast adaptation to new environments.

\subsection{Multiresolution Feature Grid}
\label{sec:implicit}
We represent the scene as learnable latent vectors anchored at the vertices of a 3D grid. These vectors act as spatial memory that is updated incrementally as new observations arrive.
Let $\calG =\{(z_i,f_i)\}_{i=1}^{M}$, where each vertex position $z_i\in \calX$ stores a latent vector $f_i\in\calF$.
For a query point $x\in \calX$, its feature is retrieved by trilinear interpolation of the eight vertex features of the voxel containing $x$,
\begin{equation}
f(x)=\sum_{i\in\mathcal{N}(x)} w(x,z_i)\,f_i,
\end{equation}
where $\mathcal{N}(x)$ is the index set of the neighboring vertices and $w(\cdot,\cdot)$ is the trilinear interpolation function.

\begin{figure}[t]
  \centering
  \includegraphics[width=0.9\linewidth]{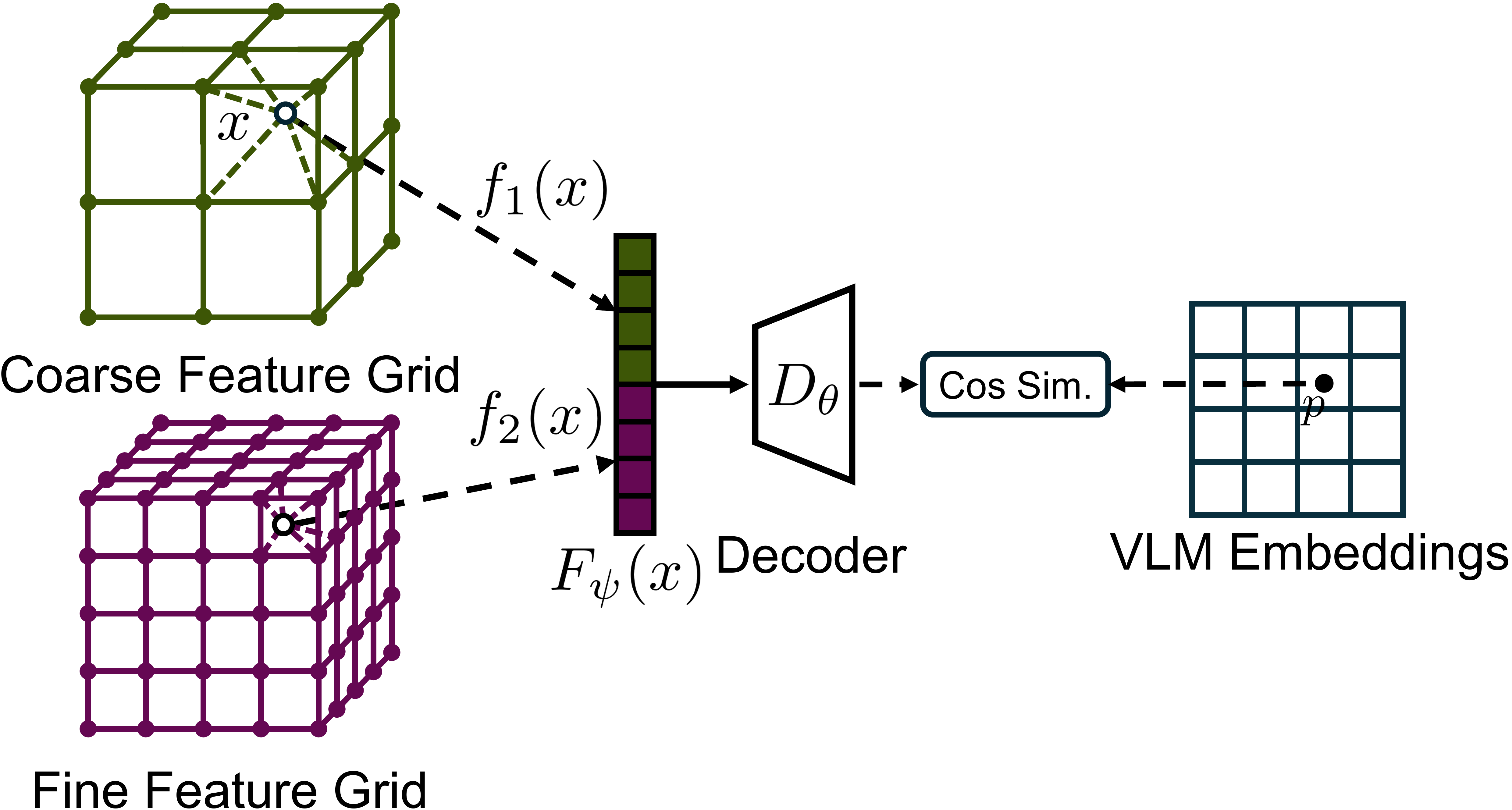}
  \caption{\textbf{Latent feature mapping.} We represent the scene with a multiresolution feature grid. For any query point $x$, we retrieve features from each level via trilinear interpolation, concatenate them to form $F_\psi(x)$, and decode with $D_\theta$ to reconstruct the target embedding. The model is trained to maximize similarity between predicted and ground-truth embeddings.}
  \label{fig:arch}
\end{figure}

To capture information at multiple scales, we use a hierarchy of $L$ grids $\{\mathcal{G}_l\}_{l=1}^{L}$, ranging from coarse ($l\!=\!1$) to fine ($l\!=\!L$) resolutions, based on the design proposed in \cite{takikawa2021neural, muller2022instant} (see \cref{fig:arch}).
Let $f_{l,i}\in\mathbb{R}^{c}$ denote the latent vector at vertex $z_{l,i}$ of grid ${\calG}_l=\{(z_{l,i}, f_{l,i})\}_{i=1}^{M_l}$, where $M_l$ is the number of vertices at level $l$. 
The collection of all latent vectors $\psi=\{\,f_{l,i}\;|\;l\!=\!1{:}L,\;i\!=\!1{:}M_l\}$ constitutes the scene-specific map parameters. 
The interpolated feature at level $l$ is $f_l(x)$.
Concatenating the features across all levels yields the final feature representation:
\begin{equation}
F_\psi(x)=\bigoplus_{l=1}^{L} f_l(x)\in \calF  \subseteq \mathbb{R}^{d}, \quad d=L\,c.
\label{eq:f_concat}
\end{equation}
where $\bigoplus$ denotes concatenation. For efficient implementation, we use a hash-based voxel grid, following~\cite{muller2022instant}. 

\subsection{Latent Map Optimization}
\label{sec:map_optimization}
The decoder $D_\theta$ maps a latent feature $F_\psi(x)$ to the target space $\calY$. The predicted feature of any query point $x\in\calX$ is
\begin{equation}
\hat{y}(x) = D_{\theta}\bigl(F_\psi(x)\bigr) \in \calY\subseteq\mathbb{R}^{k}.
\end{equation}
We implement $D_{\theta}$ as a multilayer perceptron (MLP).
The decoder is pre-trained on scenes from diverse environment configurations to learn a general mapping from the latent space~$\mathcal{F}$ to the target space~$\mathcal{Y}$.
Intuitively, $F_\psi(x)$  serves as a multiview-aggregated, compressed representation of target feature embeddings, while $D_\theta$ is trained to reconstruct them back into the target space.

During task execution, the pre-trained decoder can be kept frozen.
Adaptation to new environments is accelerated by focusing optimization on the grid parameters~$\psi$.
The parameters $\psi$ and, optionally, $\theta$ are optimized by minimizing the loss $\mathcal{L}$ in \eqref{eq:mapping}.
We align the predicted feature $\hat{y}(x)$ with the reference $y$ using the cosine distance loss, which empirically outperforms alternatives such as the $L_2$ loss.
Given a dataset $\mathcal{D}$ of coordinate-feature pairs $(x,y)$, the objective is:
\begin{equation}
\label{eq:loss_recon}
\min_{\psi,\theta}\;
\frac{1}{|\mathcal{D}|}\sum_{(x,y)\in\mathcal{D}}
\bigl[1-\,\cos\bigl(\hat{y}(x),y\bigr)\bigr].
\end{equation}
For each environment configuration (\ie, scene and object arrangement), we optimize a dedicated feature grid. 
The decoder is jointly pre-trained across all configurations for a given task.
Our mapping approach omits 3D positional encoding to avoid overfitting to absolute coordinates and to enable cross-scene generalization between different environments.
\cref{fig:arch} summarizes the overall mapping approach, instantiated for language-grounding as described in \cref{sec:latent_prob}.

\subsection{Online Latent Mapping}
\label{sec:online}
To account for changes in the environment as the task state progresses (\eg, object reconfiguration in multi-stage tasks), we update the latent map online as detailed in \cref{alg:online_mapping}. 
At the start of each policy rollout ($\tau\!=\!0$), we initialize both the feature grid and the decoder from a pre-trained offline map serving as an environment prior.
The map is updated online every $T_\text{update}$ steps using streaming robot observations. 
Here, the decoder parameters $\theta$ may be trained or frozen. We consider the frozen-decoder setting.
To preserve the consistency of the static scene, we segment out dynamic elements (\eg, robot arm) and exclude them from the updates, and we suspend updates while the robot is grasping. 
Each update performs $K_{\text{update}}$ optimization steps.
The policy is conditioned on the updated map $m_\tau$ and generates an action $a_\tau \!\sim\! \pi_\phi(\cdot \mid m_\tau, o_\tau, s_\tau, e_\ell)$.
Notably, the map parameters $\psi$ are not optimized via the policy objective in \eqref{eq:rl_bc_objectives}.

\begin{algorithm}[t!]
\caption{Online Latent Map Update}
\label{alg:online_mapping}
\begin{algorithmic}[1]
\Require pre-trained decoder $D_\theta$, learning rate $\eta$.
\State Initialize map features $\psi_0$ from a pre-trained offline map.
\For{each time step $\tau=1, 2, \dots$}
    \State Set map features $\psi_\tau \gets \psi_{\tau-1}$.
    \If{$\tau \pmod{T_\text{update}} \equiv 0$}
        \State Generate $\mathcal{D}_\tau$ from $\bigl(o_\tau, Z_\tau, (R_\tau, t_\tau)\bigr)$.
        \For{$k=1, \dots, K_\text{update}$}
            \State $\psi_\tau \gets \psi_\tau - \eta \nabla_{\psi} \mathcal{L}(\mathcal{D}_\tau; \psi_\tau)$
        \EndFor
    \EndIf
    \State $a_\tau \sim \pi_\phi(\cdot|m_\tau, o_\tau, s_\tau, e_\ell)$, where $m_\tau=(F_{\psi_\tau}, D_\theta)$.
\EndFor
\end{algorithmic}
\end{algorithm}

\subsection{Implementation Details}
\label{sec:mapping_details}
We use a two-level feature grid ($L=2$) for both room-scale and table-scale scenes, with resolutions of 0.4\,m and 0.2\,m (room) and 0.24\,m and 0.12\,m (table).
We use EVA-02-Large~\cite{fang2024eva} to obtain target VLM embeddings for supervision.
For online mapping, the map is updated every $T_\text{update}\!=\!5$ environment steps, with each update consisting of $K_\text{update}\!=\!20$ optimization steps on the map features.

\section{Map-Conditioned Policy}
\label{sec:manipulation}

This section explains how the robot manipulation policy is conditioned on the latent feature map.  
We first introduce a \emph{global map token} that captures scene-wide context from the latent map, enabling global and long-horizon reasoning (\cref{sec:map_token}). 
We then describe how the global map token is integrated into the policy, learned either with behavior cloning or reinforcement learning (\cref{sec:policy_networks}).

\subsection{Global Map Token}
\label{sec:map_token}
\begin{figure}[t]
  \centering
  \includegraphics[width=0.9\linewidth]{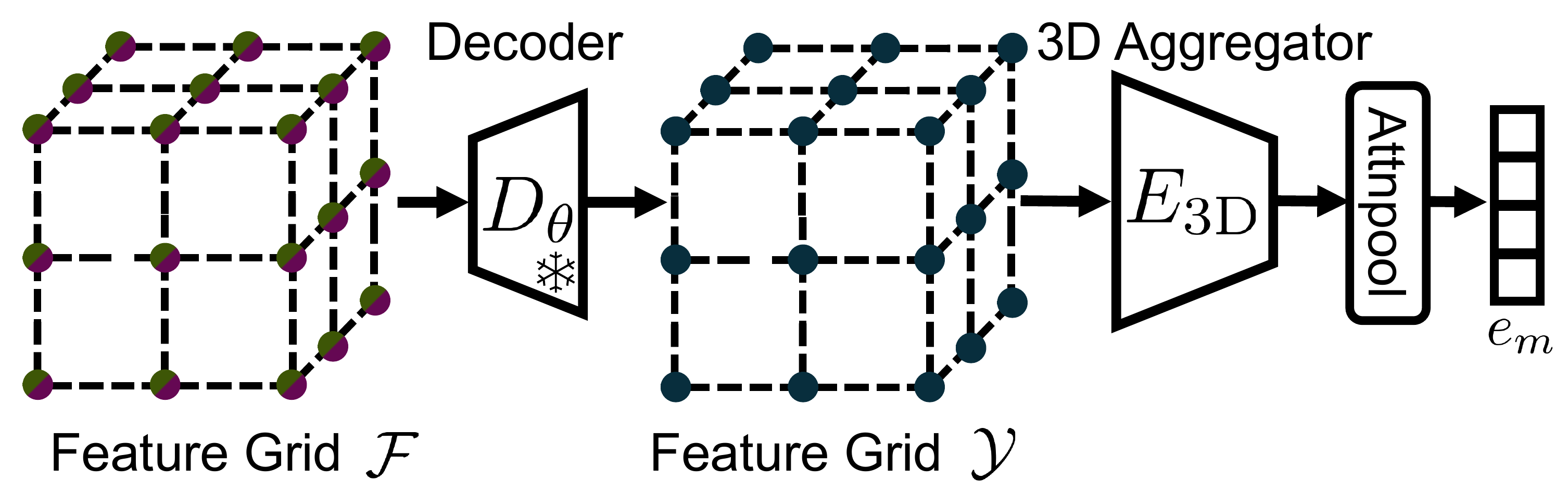}
\caption{\textbf{Global map token.} Latent features from $\mathcal{F}$ are decoded to $\mathcal{Y}$ via the decoder $D_{\theta}$ at the finest grid vertices. The 3D feature aggregator processes the coordinate-feature pairs, and its outputs are attention-pooled to produce the global map token $e_m$.}
\label{fig:map_token}
\end{figure}

The latent map enables the robot to retrieve global and long-term context about the environment beyond its immediate field of view. 
We introduce a 3D feature aggregator that distills features distributed across the scene-wide map into a compact \emph{global map token}, which then conditions the policy.
An overview of this process is shown in \cref{fig:map_token}.

First, we extract features from the map at the vertices of its finest grid level. Let $\mathcal{Z}_L=\{z_{L,i}\}_{i=1}^{M_L}$ be the coordinates of these vertices. We compute a decoded feature at each vertex using the multiresolution feature $F_\psi(z_{L,i})$ from \eqref{eq:f_concat} and the decoder $D_\theta$: $\hat{y}(z_{L,i}) = D_\theta\bigl(F_\psi(z_{L,i})\bigr)$.
We then form the set $\mathcal{S}_L$ of coordinate–feature pairs at selected vertices:
\begin{equation}
\mathcal{S}_L = \bigl\{\bigl(z_{L,i}, \hat{y}(z_{L,i})\bigr)\big|\;v_i=1, \; i{=}1,\dots,M_L\bigr\},
\end{equation}
where $v_i \in \{0, 1\}$ is a binary mask selecting task-relevant vertices (\eg, excluding unoccupied or background regions).
These are fed into a 3D feature aggregator, $E_{\mathrm{3D}}$, and its outputs are attention-pooled to produce a global map token:
\begin{equation} \label{eq:map_token}
 e_m = \operatorname{Attn-Pool}\Bigl(E_{\mathrm{3D}}(\mathcal{S}_L)\Bigr).
\end{equation}
The token $e_m$ is used to condition the policy $\pi_\phi$.
With an offline map, the global map token is time-invariant; with online mapping (\cref{sec:online}), it depends on the time step~$\tau$ as the task state changes.
The parameters of $E_{\mathrm{3D}}$ are included in $\phi$ and jointly optimized via the objective in \eqref{eq:rl_bc_objectives}.

The $E_{\mathrm{3D}}$ architecture matches the environment's scale. 
For large room-scale environments, we use a Point Transformer~\cite{zhao2021point}, whose hierarchical attention captures long-range spatial relationships. 
For smaller tabletop scenes, we utilize a lightweight PointNet~\cite{qi2017pointnet}, whose simpler structure is sufficient for compact environments and its computational efficiency is advantageous for sample-intensive RL. We encode coordinates with sinusoidal positional encodings~\cite{mildenhall2021nerf}, concatenate them with features, and input them to PointNet.

\subsection{Map-Conditioned Policy Network}
\label{sec:policy_networks}

\begin{figure}
    \centering
    \includegraphics[width=0.9\linewidth]{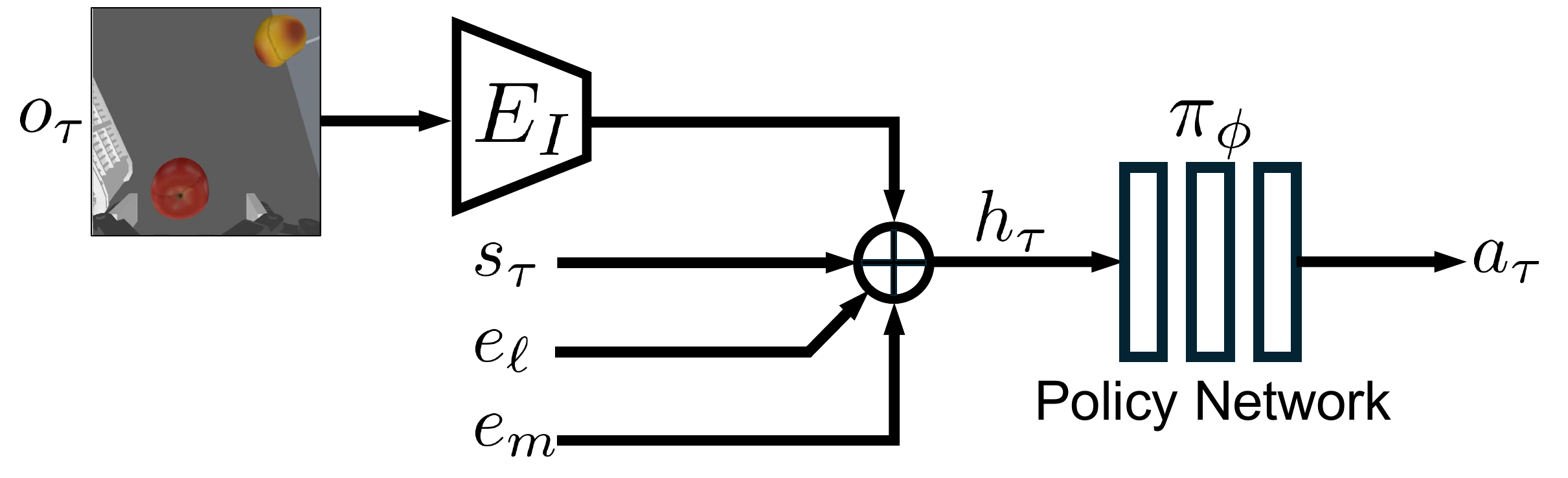}
    \caption{\textbf{Map-conditioned policy network.} Proprioceptive state $s_\tau$, image features $E_I(o_\tau)$, task embedding $e_\ell$, and global map token $e_m$ are concatenated to form a joint embedding $h_\tau$, which is mapped to an action $a_\tau$ by the policy network $\pi_{\phi}$. }
  \label{fig:map_policy}
\end{figure}

We treat the map token $e_m$ in \eqref{eq:map_token} as an additional state input to the policy networks. 
We first outline a generic integration scheme and, then, specify its instantiation for BC and RL.
The policy inputs are image features $E_{I}(o_\tau)$, proprioceptive state $s_\tau$, task embedding $e_\ell$, and the global map token $e_m$. Concatenating these inputs yields a joint embedding $h_\tau$:
\begin{equation}
h_\tau = \operatorname{Concat}\bigl(E_{I}(o_\tau), s_\tau, e_\ell, e_m\bigr),
\end{equation}
which is provided as input to the policy network $\pi_\phi$ to produce an action $a_\tau$.
The action generation and concatenation details depend on the policy architecture and learning method. The overall architecture is illustrated in \cref{fig:map_policy}.

\paragraph{Map-Conditioned BC}
For BC, we formulate the policy architecture using ACT~\cite{zhao2023learning}. 
Image features from a Transformer encoder are concatenated channel-wise with the proprioceptive state, task embedding, and global map token to form the joint embedding $h_\tau$. 
Then, $h_\tau$ is fed into a Transformer decoder, which predicts an action sequence $a_{\tau:\tau+H-1}$. 
The task embedding $e_\ell$ is obtained by encoding a natural language command (\eg, ``pick up the bowl'') using a VLM text encoder~\cite{fang2024eva}. 
Unlike the original ACT, we use an exponentially decaying loss weight on future action steps, prioritizing accuracy on immediate actions.
This change is empirically effective for our expert demonstrations, which are produced by an RL policy and less smooth than human demonstrations.
At test time, only the first action of each predicted chunk is executed.
Our BC policy uses DINOv2-ViT-S~\cite{oquab2023dinov2} as the visual backbone $E_{I}$. The Transformer has a 4-layer encoder and 6-layer decoder, predicting action sequences over a 16-step horizon ($H\!=\!16$).

\begin{figure*}[t]
  \centering
  \includegraphics[width=\linewidth]{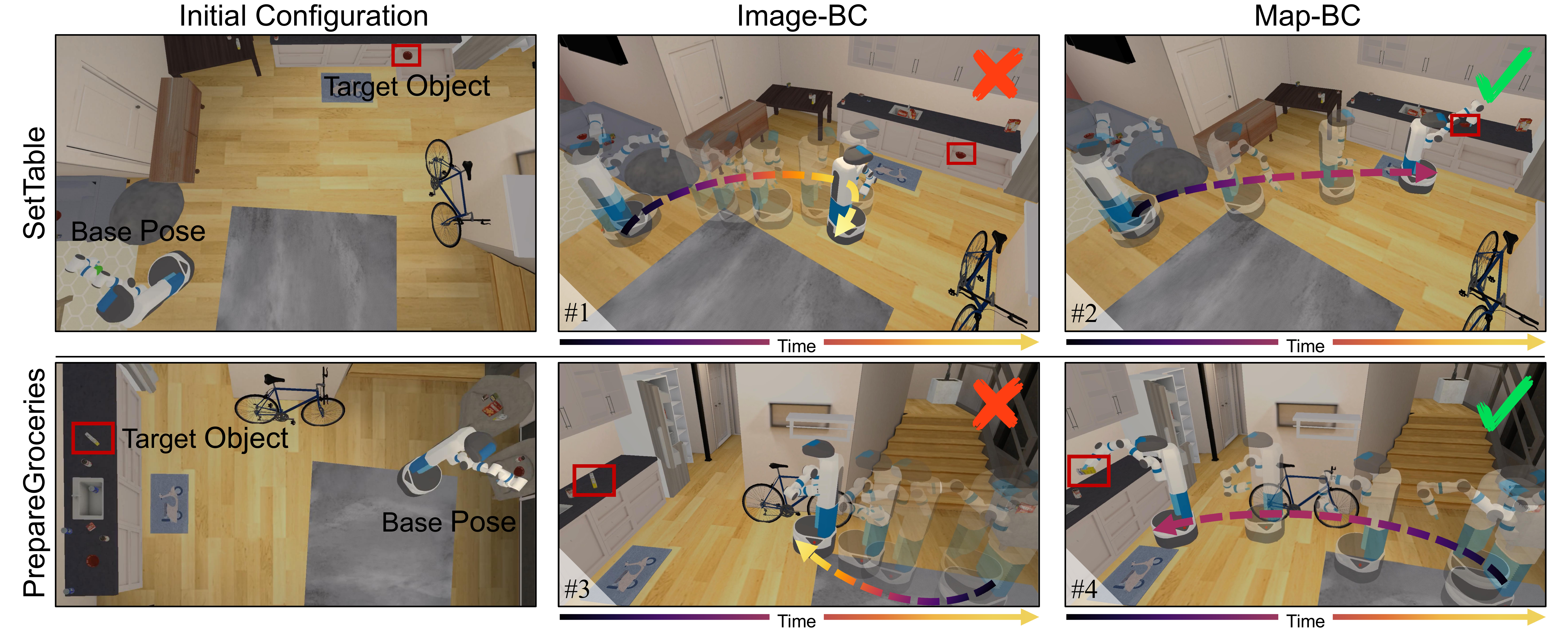} %
  \caption{Qualitative comparison on home-rearrangement tasks under out-of-distribution conditions. At the start of each episode, the robot is placed at a distant base pose unseen during training, with the target object completely outside of the robot's current field of view. Image-BC (\#1 and \#3) fails to localize the object in these settings, resulting in inefficient trajectories that fail to reach the target. In contrast, Map-BC (\#2 and \#4) successfully navigates to and grasps the object, completing the task with direct and efficient trajectories.}
  \label{fig:qualitative} 
\end{figure*}

\begin{table*}[t]
\centering
\caption{Performance of BC policies using different visual representations for an object-picking task from three home-rearrangement benchmarks.  
For each benchmark, we report success rate (SR↑) and episode reward (ER↑) on both the training scene (ID) and the novel scene (OOD), averaged over three runs.  
Best and second-best results are highlighted in \textbf{bold} and \underline{underlined}, respectively.}
\label{tab:baseline}

\renewcommand{\arraystretch}{1.2}

\resizebox{\linewidth}{!}{%
\begin{tabular}{|l|cc|cc|cc|cc|cc|cc|}
\hline
& \multicolumn{4}{c|}{TidyHouse-Pick}
& \multicolumn{4}{c|}{PrepareGroceries-Pick} & \multicolumn{4}{c|}{SetTable-Pick} \\
\hline
Method & SR (ID) & SR (OOD) & ER (ID) & ER (OOD) & SR (ID) & SR (OOD) & ER (ID) & ER (OOD) & SR (ID) & SR (OOD) & ER (ID) & ER (OOD) \\
\hline
Image-BC \cite{zhao2023learning}                        & \underline{0.31} & 0.29 & \underline{0.50} & 0.50 & 0.30 & 0.25 & 0.46 & \underline{0.44} & \underline{0.55} & 0.49 & 0.58 & 0.56 \\
Uplifted~\cite{ke20243d}         & 0.30 & \underline{0.30} & 0.48 & \underline{0.50} & \underline{0.31} & \underline{0.28} & \textbf{0.48} & 0.44 & \textbf{0.56} & \underline{0.51} & \textbf{0.59} & \underline{0.57} \\
Point Cloud~\cite{ze20243d}      & 0.15 & 0.13 & 0.42 & 0.41 & 0.16 & 0.16 & 0.38 & 0.37 & 0.41 & 0.32 & 0.49 & 0.45 \\
\rowcolor{gray!20} Map-BC (offline)                    & \textbf{0.33} & \textbf{0.31} & \textbf{0.53} & \textbf{0.52} & \textbf{0.33} & \textbf{0.30} & \underline{0.47} & \textbf{0.46} & 0.55 & \textbf{0.54} & \underline{0.59} & \textbf{0.60} \\
\hline
\end{tabular}%
}
\vspace*{-2ex}
\end{table*}

\paragraph{Map-Conditioned RL}
For RL, we use PPO~\cite{schulman2017proximal} with an actor-critic architecture where both heads are two-layer MLPs. 
The actor head outputs an action distribution $\pi_\phi(\cdot \mid\!h_\tau)$, while the critic head estimates the value $V(h_\tau)$, used to compute the advantage in the PPO objective. 
Both heads are conditioned on the joint embedding $h_\tau$, formed via feature-wise concatenation of the proprioceptive state, the map token, a learnable task embedding $e_\ell$ specifying the target object, and image features flattened then projected by an MLP.
We use a frozen DINOv2-ViT-S~\cite{oquab2023dinov2} whose output is processed by a two-layer MLP as the visual backbone $E_{I}$.
To improve sample efficiency, we adopt a two-stage curriculum. 
First, we pre-train a map-agnostic, image-based policy by replacing the global map token with a zero vector. 
We then fine-tune the policy with the map token enabled via a learnable gating mechanism. 
A trainable sigmoid gate applies element-wise scaling to the map token, allowing the policy to gradually incorporate map features during fine-tuning.

\section{Evaluation}
\label{sec:evaluation}

We evaluate \AlgName on its ability to (1) reason globally, particularly for targets beyond the field of view, and (2) handle multi-stage tasks requiring long-term context.
We test \AlgName in two settings: a home-rearrangement mobile manipulation task (\cref{sec:mobile_manipulation}) and a sequential pick-and-place task (\cref{sec:sequential_manipulation}). For the latter, we also demonstrate zero-shot sim-to-real transfer of our learned policy to a physical robot.
In both settings, \AlgName demonstrates superior performance over policies that rely solely on image-based reasoning.

Across all experiments, performance is evaluated using success rate (SR) and episode reward (ER). ER comprises object reach shaping, a grasp bonus, and a success bonus. We report both in-distribution (ID) and out-of-distribution (OOD) results. All experiments run in the ManiSkill simulator~\cite{tao2024maniskill3}.

\subsection{Mobile Manipulation Experiment}
\label{sec:mobile_manipulation}

\myParagraph{Setup}
Our setup is based on Pick Subtasks from the ManiSkill-HAB benchmark~\cite{shukla2024maniskill, szot2021habitat}.
Unlike the original benchmarks, we train our policy on demonstrations from only a single scene (\texttt{sc1-13}), which are generated using the RL policy from~\cite{shukla2024maniskill}. 
We evaluate on the training scene (ID, \texttt{sc1-13}) and a novel, unseen scene with a different layout and object arrangement (OOD, \texttt{sc1-10}).
To test generalization with a map, OOD evaluation uses a pre-generated latent map of the novel scene, without using any expert demonstrations from the scene.

We compare three baseline policies that use different visual representations and our map-conditioned BC policy.
\begin{itemize}[leftmargin=1.3em, itemsep=0pt, topsep=2pt]
    \item Image-BC: Image-based policy (ACT \cite{zhao2023learning}).
    \item Uplifted: Image-based policy whose features are lifted into transient 3D tokens via 3D RoPE \cite{ke20243d, gervet2023act3d}.
    \item Point Cloud: Point cloud-based policy with point cloud observations processed by a 3D feature encoder \cite{qi2017pointnet, ze20243d}.
    \item Map-BC (offline): The proposed map-conditioned policy that uses a pre-generated offline map.
\end{itemize}
To ensure a fair comparison, all methods use the same policy architecture and hyperparameters.
  
\myParagraph{Results}
\cref{tab:baseline} shows that Map-BC outperforms the baselines on most benchmarks, with especially large gains on challenging tasks such as TidyHouse, which includes nine target objects.
We attribute this to stronger scene understanding, enabling efficient target localization and navigation.

\cref{fig:qualitative} shows two illustrative runs comparing Map-BC and Image-BC policies in mobile manipulation tasks that require global reasoning.  In the original dataset, localization is simplified by initializing the robot near the target and orienting it towards the target.
Instead, we apply substantial translational and rotational perturbations to the robot's initial pose, resulting in the target being far outside the robot's initial field of view.
The Image-BC baseline produces erratic and inefficient trajectories and fails to reach the target, whereas Map-BC drives toward the target object and completes the task.
These results indicate that the latent map enables the policy to perform global reasoning over the scene.

\begin{figure}[t]
    \centering
    \includegraphics[width=\linewidth]{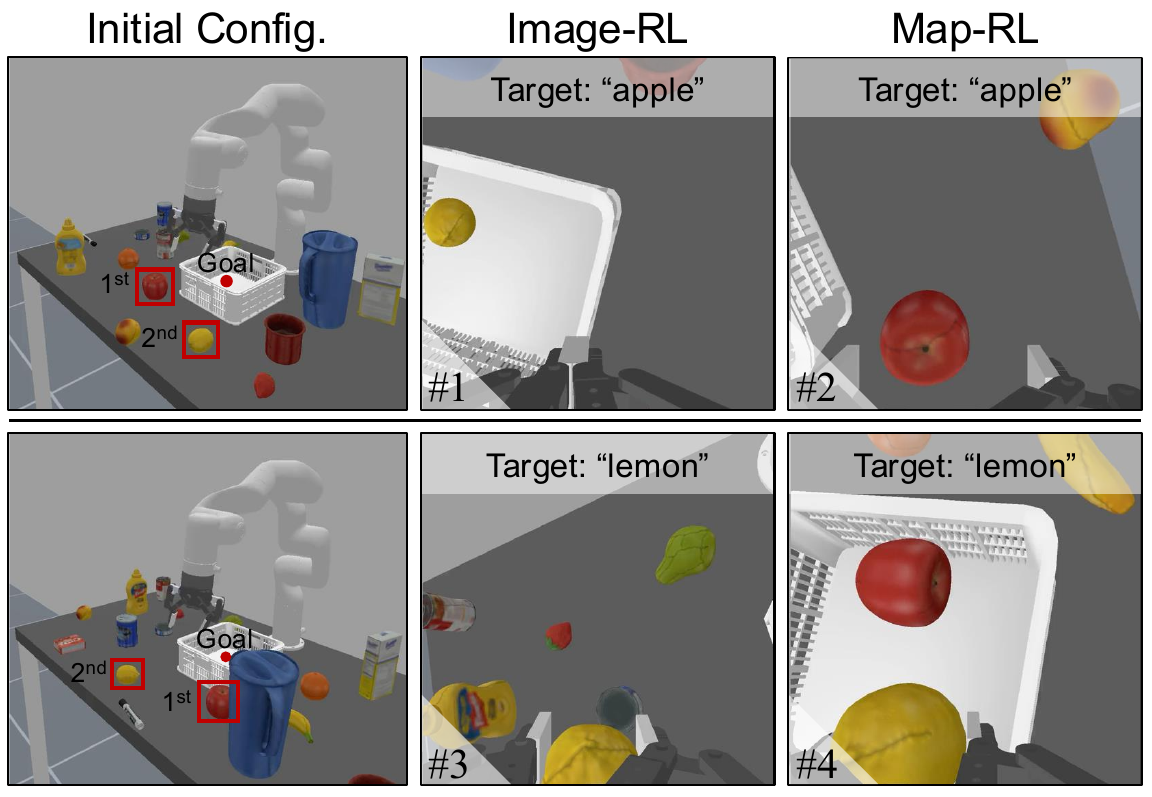}
    \caption{Qualitative results from sequential manipulation using only egocentric views. Image-RL fails to localize the second object (\#1) or the goal (\#3) once they are out of view. In contrast, Map-RL leverages the map as spatial memory, locating the object and goal and completing the task sequence (\#2 and \#4).}
    \label{fig:tabletop}
\end{figure}

\begin{table}[t]
\centering
\caption{Performance of RL policies on the sequential pick-and-place task. We evaluate on 100 training (ID) and 20 novel (OOD) scenes. We report success rate (SR↑) and episode reward (ER↑), averaged over three runs. Best results are highlighted in \textbf{bold}.}
\label{tab:tabletop}
\renewcommand{\arraystretch}{1.2}
\resizebox{0.9\columnwidth}{!}{%
\begin{tabular}{|l|cc|cc|}
\hline
& \multicolumn{4}{c|}{Sequential Pick-and-Place} \\
\hline
Method & SR (ID) & SR (OOD) & ER (ID) & ER (OOD) \\
\hline
Image-RL \cite{schulman2017proximal}               & 0.82          & 0.75          & 0.77          & 0.7         \\
\rowcolor{gray!20} Map-RL (offline)       & 0.94          & 0.95          & 0.85          & 0.84         \\
\rowcolor{gray!20} Map-RL (online)         & \textbf{0.97}          & \textbf{1.00}          & \textbf{0.87}          & \textbf{0.88}         \\
\hline
\end{tabular}%
}
\vspace*{-2ex}
\end{table}

\subsection{Sequential Manipulation Experiment}
\label{sec:sequential_manipulation}

\myParagraph{Setup}
We design a two-stage sequential pick-and-place task to evaluate whether the latent map improves long-horizon manipulation.
The task requires picking objects from a cluttered tabletop and placing them in a basket in a prescribed order (\cref{fig:tabletop}). 
For each episode, we sample an environment from a predefined set of scenes that differ in the arrangements and poses of clutter and target objects. 
The target set is fixed across episodes, and the execution order of targets is randomized per episode.
To mimic limited visibility in mobile manipulation, the robot relies solely on an egocentric camera (\ie, no global view of the scene). 
The scenes contain occlusions, making it challenging to localize the target objects.
Experiments are conducted on a uFactory xArm6 robot.
We evaluate on 100 training (ID) scenes and 20 novel (OOD) scenes with unseen object arrangements.

We compare an image-based baseline with two variants of the map-conditioned RL policy.
\begin{itemize}[leftmargin=1.3em, itemsep=0pt, topsep=2pt]
    \item Image-RL: Image-based policy.
    \item Map-RL (offline): The proposed map-conditioned policy that uses a pre-generated offline map.  
    \item Map-RL (online): The proposed map-conditioned policy that uses an online-updated map, as described in \cref{sec:online}.
\end{itemize}
All policies are trained with PPO~\cite{schulman2017proximal} under the same policy architecture and hyperparameters.

\myParagraph{Results}
The results of the sequential manipulation experiment are summarized in \cref{tab:tabletop} and illustrated in \cref{fig:tabletop}. 
Map-RL (offline) and Map-RL (online) both outperform Image-RL in SR and ER, with a larger margin in OOD settings.
In \cref{fig:tabletop}, Image-RL fails to localize the second object or the goal once they leave the egocentric view. In contrast, our Map-RL policy leverages the latent map as spatial memory, enabling it to locate the objects and complete the task sequence.
Furthermore, Map-RL (online) shows an advantage over Map-RL (offline). The latter relies on an offline map capturing only the initial object arrangement, whereas the former's online updates provide temporal memory that allows the policy to track the task state.

\begin{figure}[t]
    \centering
    \includegraphics[width=1.0\linewidth]{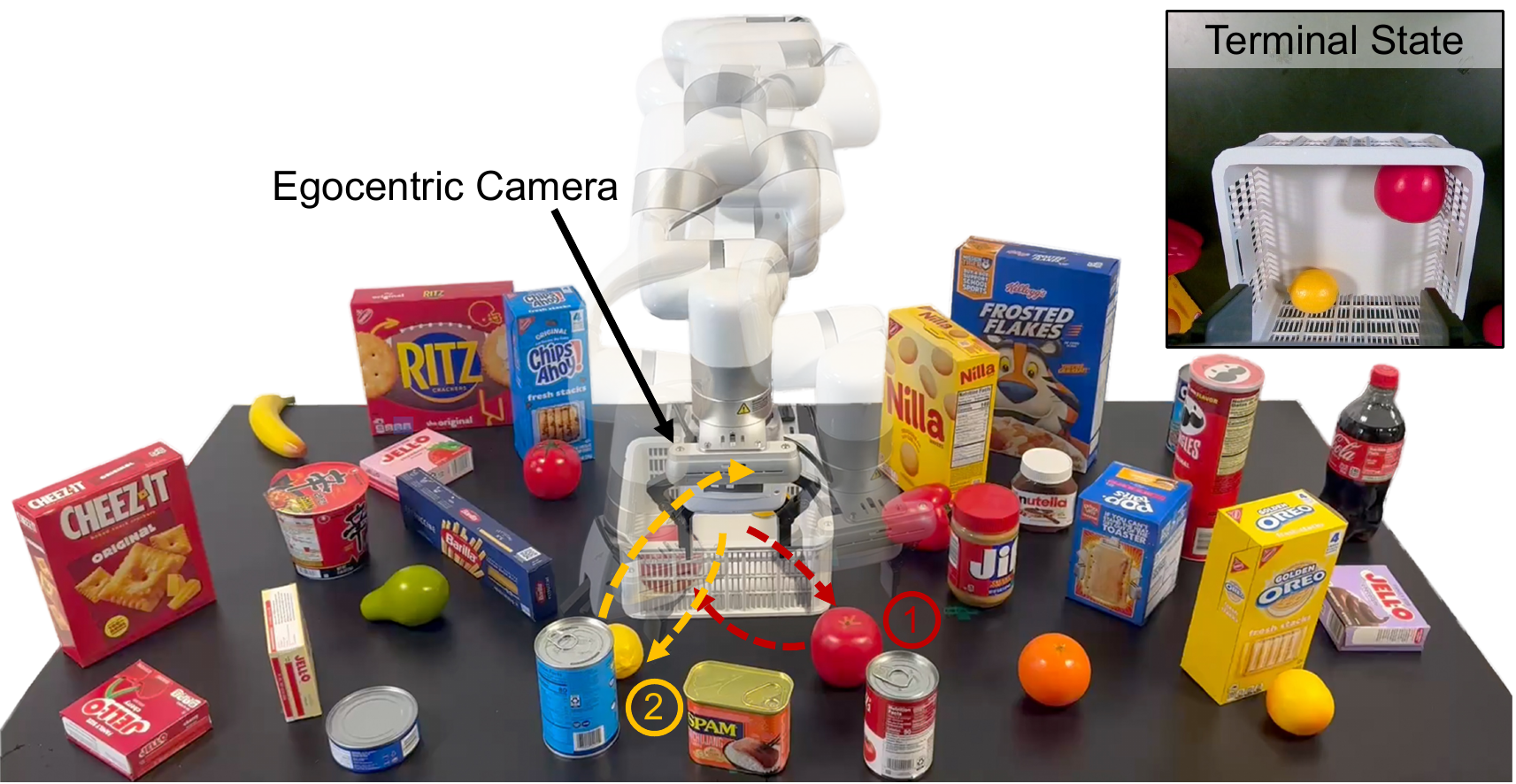}
    \caption{Real-world setup mirroring the sequential manipulation task from simulation. Our policy is transferred from simulation to the real robot in a zero-shot manner and completes the task.}
    \label{fig:real_robot}
    \vspace*{-2ex}
\end{figure}

\myParagraph{Real robot deployment}
We deploy the Map-RL (offline) policy, trained in simulation, on a uFactory xArm6 robot in a zero-shot manner. 
Our real-world setup, shown in \cref{fig:real_robot}, closely mirrors the simulation environment with aligned coordinate frames, identical objects (\eg, table, basket, and targets), and matched egocentric camera pose.
We first build an offline latent map of the real scene from a sequence of egocentric RGB-D images and camera poses estimated from the robot's forward kinematics.
We then deploy the policy using this map as input.
We do not use additional sim-to-real transfer techniques; instead, we rely on a frozen DINOv2 visual backbone that is robust to the visual domain gap.
The policy successfully completes the sequential manipulation task on the real robot.
The sim-to-real gap still makes grasping challenging in scenes with distant targets.
\cref{fig:real_robot} shows snapshots from the policy rollout.

\section{Conclusion}
\label{sec:conclusion}
While 3D maps have long been core components of robot navigation, they have largely been overlooked in learning manipulation policies.
In this paper, we presented a 3D latent map formulation that offers key advantages for manipulation: (i) perception beyond the robot's field of view and (ii) observation aggregation over long horizons.
Building on these advantages, we proposed an end-to-end approach that couples a 3D latent map with mobile manipulation policy learning, providing the robot with extended spatial and temporal context.
The experiments demonstrate that the map-conditioned policy is able to reason globally, leveraging the map as spatiotemporal memory for scene-level mobile manipulation and sequential manipulation tasks.

Several avenues for future work are possible.
First, while our latent feature map provides global context, the policy still relies on local image features. 
Future work could reduce this dependency by developing dynamic scene representations to capture the motion of the robot and other objects.
Second, the map-conditioned RL policy is trained using an on-policy method, which can be sample-inefficient, requiring pre-training from an image-based policy. 
Integrating our method with off-policy or model-based RL could mitigate these problems. Third, our map-conditioned policy uses Point Transformer~\cite{zhao2021point} for behavior cloning but a lighter PointNet~\cite{qi2017pointnet} for RL, as the former is too expensive for online training. Developing a unified and efficient 3D aggregation model would enable online policy training. Finally, evaluating our approach on larger-scale, longer-horizon mobile manipulation tasks is needed to further validate its effectiveness. Extending to more complex dexterous manipulation scenarios could also demonstrate its broader applicability.


\bibliographystyle{IEEEtran}
\bibliography{references}


\end{document}